\documentclass[sigconf]{acmart}

\AtBeginDocument{%
  \providecommand\BibTeX{{%
    \normalfont B\kern-0.5em{\scshape i\kern-0.25em b}\kern-0.8em\TeX}}}

\usepackage[utf8]{inputenc}
\usepackage{amsmath,amsfonts}
\usepackage{algorithm} 
\usepackage{algorithmic}
\usepackage{graphicx}
\usepackage{textcomp}
\usepackage{url}
\usepackage{xcolor}
\usepackage{multirow}
\usepackage{tikz}
\usepackage{mathrsfs}
\usepackage{enumitem}
\usepackage{listings}
\usepackage{footmisc}
\usepackage{courier}
\usepackage{subfig} 
\usepackage{hyperref}
\usepackage{xspace}
\usepackage{multirow}
\usetikzlibrary{arrows.meta, positioning, shapes.multipart}

\def\@onedot{\ifx\@let@token.\else.\null\fi\xspace}

\newcommand{\dataset}{\textit{UncertaintyZoo}\xspace}

\setcopyright{none}

\begin{document}

\title{\dataset: A Unified Toolkit for Quantifying Predictive Uncertainty in Deep Learning Systems}

\title{\dataset: An Uncertainty Quantification Toolkit for Large Language Models}

\author{Xianzong Wu}

\affiliation{
 \institution{Tianjin University}
 \country{China}
 }

 \author{Xiaohong Li}

\affiliation{
 \institution{Tianjin University}
 \country{China}
 }

\author{Lili Quan}

\affiliation{
 \institution{Tianjin University}
 \country{China}
 }

\author{Qiang Hu}

\affiliation{
 \institution{Tianjin University}
 \country{China}
 }

\begin{abstract}

Large language models~(LLMs) are increasingly expanding their real-world applications across domains, e.g., question answering, autonomous driving, and automatic software development. Despite this achievement, LLMs, as data-driven systems, often make incorrect predictions, which can lead to potential losses in safety-critical scenarios. To address this issue and measure the confidence of model outputs, multiple uncertainty quantification~(UQ) criteria have been proposed. However, even though important, there are limited tools to integrate these methods, hindering the practical usage of UQ methods and future research in this domain. To bridge this gap, in this paper, we introduce \dataset, a unified toolkit that integrates 29 uncertainty quantification methods, covering five major categories under a standardized interface. Using \dataset, we evaluate the usefulness of existing uncertainty quantification methods under the code vulnerability detection task on CodeBERT and ChatGLM3 models. The results demonstrate that \dataset effectively reveals prediction uncertainty. The tool with a demonstration video is available on the project site \url{https://github.com/Paddingbuta/UncertaintyZoo}.



\end{abstract}

\maketitle

\section{Introduction}

Large Language Models~(LLMs) have demonstrated remarkable success across a broad range of applications, including security-critical domains such as autonomous driving, software vulnerability detection, and threat intelligence. However, despite their strong performance, LLMs remain prone to erroneous predictions, particularly when encountering out-of-distribution~(OOD) or ambiguous inputs where model confidence is low~\cite{gal2016dropout}. In safety-sensitive scenarios, such errors can lead to severe security and reliability risks. Therefore, prediction interpretation is crucial, and there is a need to quantify the prediction confidence.

To do so, Uncertainty Quantification~(UQ) has emerged as a critical technique to estimate the confidence of LLM output, providing essential information on when predictions can or cannot be trusted. Prior studies introduce a broad spectrum of UQ methods for LLMs, ranging from Monte Carlo Dropout and mutual information to self-consistency sampling~\cite{xiao2019quantifying}, aiming to assess model confidence and improve prediction robustness under distribution shifts or ambiguous inputs. Several surveys~\cite{abdar2021review, huang2024survey} categorize such techniques from a theoretical perspective. However, few tools integrate these UQ methods in a unified framework, limiting their practical adoption in real-world systems and hindering systematic comparisons and reproducibility.

A unified UQ framework is thus essential to advance trustworthy and uncertainty-aware LLM applications. Such a framework (1) offers a standardized interface to efficiently apply diverse UQ methods for confidence estimation, reducing the overhead of fragmented or ad-hoc implementations; (2) enables systematic and fair comparisons across alternative techniques under consistent experimental settings, fostering deeper insights into their practical strengths and limitations; and (3) provides a common, extensible platform to facilitate the development and evaluation of novel UQ approaches, thereby promoting progress in this field. However, current toolkits~\cite{nikitin2025luq,bouchard2025uq,bouchard2025uqlm} cover only a narrow subset of UQ methods, in both quantity and category, making comprehensive, systematic evaluation infeasible. 

To address this gap, we introduce \dataset, a unified and plugin-oriented UQ framework that integrates 29 state-of-the-art UQ techniques into a single coherent platform for LLMs. To achieve this, we first categorize existing UQ methods into five representative families based on their underlying principles: \textit{predictive distribution methods, ensemble-based methods, input-level sensitivity methods, reasoning-level analyses, and representation-based methods}. We then implement methods from each category under a standardized interface, enabling seamless integration of additional techniques and promoting consistency across use cases.

\dataset supports both discriminative and generative models and is designed for practical integration into LLM pipelines. To demonstrate its utility, we apply \dataset to the code vulnerability detection task on CodeBERT and ChatGLM3 models. Experimental results show that 1) token-probability-based UQ methods are relatively more effective than other methods on our considered tasks; 2) existing UQ methods cannot precisely quantify the prediction uncertainty of generative models, highlighting the need for new UQ methods.

In summary, this paper makes the following contributions:

\begin{itemize}[leftmargin=*]
\item We propose \dataset, a generic and comprehensive framework covering 29 representative UQ methods to support confidence estimation in LLM.

\item We evaluate \dataset on the code vulnerability detection task~(Devign) and find that token-probability-based UQ methods are more effective than others and there is a need for designing new UQ methods for generative models and tasks.



\end{itemize}


\section{The Framework of \dataset}

\begin{figure}[h]
\centering
\begin{tikzpicture}[
    node distance=0.8cm and 1cm,
    every node/.style={font=\small},
    box/.style={draw, rounded corners, minimum width=3.6cm, minimum height=0.8cm, align=center},
    catbox/.style={draw, thick, rounded corners, minimum width=5.5cm, minimum height=1.4cm, align=center, fill=blue!5},
    arrow/.style={-{Latex}, thick}
]

\node[box, fill=gray!10] (input) {Input Sample \\ + Model + Config};

\node[box, below=of input, fill=blue!10] (quantifier) {Quantifier Engine};

\node[catbox, below=of quantifier] (uqmethods) {
\textbf{Uncertainty Method Categories} \\
A: Predictive \quad B: Ensemble \quad C: Input-Level \\
D: Reasoning \quad E: Representation
};

\node[box, below=of uqmethods, fill=green!10] (output) {Uncertainty Score(s) \\ + Visualization};

\draw[arrow] (input) -- (quantifier);
\draw[arrow] (quantifier) -- (uqmethods);
\draw[arrow] (uqmethods) -- (output);

\end{tikzpicture}
\caption{Workflow of the \dataset Toolkit}
\label{fig:uqzoo_workflow}
\end{figure}

Fig.~\ref{fig:uqzoo_workflow} illustrates the workflow of \dataset. Given an input (e.g., code snippets) and the code model (discriminative or generative), \dataset uses the code model to generate predictions or reasoning chains. 
It then dispatches the computation to one of five method categories via a unified interface—predictive distribution, ensemble, input-level sampling, reasoning-aware, and representation-based. Each category exposes a standard API, and \dataset invokes the corresponding routine (e.g., using output probabilities, multiple stochastic samples, or semantic/topological analyses of reasoning trajectories) to produce a scalar uncertainty score. 





\textit{Comparison between different UQ frameworks.} 
Table~\ref{tab:dataset} compares existing UQ frameworks with \dataset. Compare to existing tools, \dataset has two key advantages: 1) \dataset covers the most unique UQ methods~(29), significantly outperforming other tools~(e.g., 12 for \cite{bouchard2025uqlm}). 2) \dataset supports more diverse UQ categories. Only \dataset covers all five UQ categories, other tools consider at most three categories, demonstrating the comprehensiveness of our toolkit.

\begin{table}[]
\centering
\caption{Comparison between different uncertainty quantification frameworks.}
\label{tab:dataset}
\small  
\begin{tabular}{c|c|c}
\hline
\textbf{Benchmark} & \textbf{Categories} & \textbf{\# Methods} \\ \hline
Nikitin~\emph{et al.}~\cite{nikitin2025luq} & Predictive & 6  \\ \hline
Bouchard~\emph{et al.}~\cite{bouchard2025uq} & Predictive, Ensemble, Reasoning & 9  \\ \hline
Bouchard~\emph{et al.}~\cite{bouchard2025uqlm} & Predictive, Ensemble  & 12  \\ \hline
\dataset~(Ours) & All & 29  \\ \hline
\end{tabular}
\end{table}

\subsection{Uncertainty Quantification Methods.}


Table \ref{tab:UCmethods} summarizes all 29 methods along with their category, task type, and processing level (e.g., token, input, or reasoning level). 

\begin{table}[!t]
\centering
\caption{Uncertainty quantification methods in \dataset. C: Classification. NC: Non-Classification}
\label{tab:UCmethods}
\resizebox{0.5\textwidth}{!}{
\begin{tabular}{|c|l|c|c|}
\hline
\textbf{Category} & \textbf{Method Name} & \textbf{Task Type} & \textbf{Level} \\ \hline
\multirow{11}{*}{\begin{tabular}[c]{@{}c@{}}Predictive \\ Distribution\end{tabular}} & Average Negative Log-Likelihood & C & Token \\
 & Average Probability & C & Token \\
 & Perplexity & C & Token \\
 & Maximum Token Entropy & C & Token \\
 & Average Prediction Entropy & C & Token \\
 & Token Impossibility Score & C & Token \\
 & Margin Score & C & Output \\
 & Maximum Probability & C & Output \\
 & Least Confidence & C & Output \\
 & Predictive Entropy & C & Output \\
 & DeepGini & C & Output \\ \hline
\multirow{9}{*}{Ensemble} & Expected Entropy & C & Output \\
 & Mutual Information (BALD) & C & Output \\
 & Monte Carlo Dropout Variance & C & Output \\
 & Class Prediction Variance & C & Output \\
 & Class Probability Variance & C & Output \\
 & Sample Variance & C & Output \\
 & Maximum Difference Variance & C & Output \\
 & Minimum Variance & C & Output \\
 & Cosine Similarity of Embeddings & C & Embedding \\ \hline
\multirow{3}{*}{\begin{tabular}[c]{@{}c@{}}Input-Level \\ Sampling\end{tabular}} & Self-Perturbation Uncertainty Quantification & NC & Input \\
 & In-Context Learning Sampling & NC & Input \\
 & Input Clarification Ensembles & NC & Input \\ \hline
\multirow{5}{*}{Reasoning} & Uncertainty-Aware Attention Gradients (UAG) & NC & Reasoning \\
 & Chain-of-Thought Uncertainty (CoT-UQ) & NC & Reasoning \\
 & Tree-of-Thought Uncertainty (TouT) & NC & Reasoning \\
 & Topology-Based Uncertainty (TopologyUQ) & Both & Reasoning \\
 & Stable Explanation Confidence & Both & Reasoning \\ \hline
Representation & Logit Lens Entropy & C & Hidden-State \\ \hline
\end{tabular}
} 
\end{table}

\subsubsection{Predictive Distribution Methods} estimate model confidence by analyzing the probability distribution over the model's output. They consider either the overall distribution over classes or token-level probabilities in sequence models, making them lightweight, task-agnostic, and widely applicable.



\begin{itemize}[leftmargin=*]
    \item \textbf{Maximum Probability} — takes the highest predicted class probability as a proxy for confidence. Lower values imply the model is more uncertain.

    \item \textbf{Least Confidence} — defined as $1 - \max(p(y|x))$, it captures uncertainty when the model lacks a dominant prediction.

    \item \textbf{Margin Score} — computes the difference between the top two predicted class probabilities. Smaller margins suggest ambiguous predictions.

    \item \textbf{Predictive Entropy} — measures the overall uncertainty over class distributions:
    \[
    H(y|x) = -\sum_{i=1}^{C} p(y_i|x) \log p(y_i|x)
    \]
    It reflects the dispersion of the predicted probability mass across all classes.

    \item \textbf{Average Negative Log-Likelihood}, \textbf{Average Probability}, and \textbf{Perplexity} — evaluate how well the predicted probabilities align with ground-truth labels. These are standard metrics in classification and language modeling.

    \item \textbf{Maximum Token Entropy} — identifies the most uncertain token position in a sequence by selecting the maximum entropy across all tokens.

    \item \textbf{Average Prediction Entropy} — computes the average token-level entropy, capturing overall uncertainty throughout the sequence.

    \item \textbf{Token Impossibility Score} — detects the most improbable decision by analyzing the least likely token across the prediction sequence.

    \item \textbf{DeepGini} — derived from the Gini index, it estimates uncertainty based on how uniform the class probabilities are:
    \[
    \mathit{DeepGini} = 1 - \sum_{i=1}^{C} p(y_i|x)^2
    \]
    A more uniform distribution leads to higher DeepGini values, indicating greater uncertainty.
\end{itemize}

\subsubsection{Ensemble-Based Methods} estimate uncertainty by analyzing the variability of predictions across multiple stochastic runs (e.g., Monte Carlo Dropout) or different model instances. They quantify uncertainty by assessing disagreement among predictions, leveraging variance in class probabilities, predicted labels, or semantic embeddings. Representative methods in this category include:


\begin{itemize}[leftmargin=*]
    \item \textbf{Expected Entropy} — computes the average entropy over $S$ sampled predictions:
    \[
    \mathit{H}_{\text{expected}} = \frac{1}{S} \sum_{s=1}^{S} \left[ -\sum_{i=1}^{C} p(y_i | x_s) \log p(y_i | x_s) \right]
    \]
    It captures the aleatoric uncertainty by averaging the randomness in each sampled output.

    \item \textbf{Mutual Information (BALD)} — quantifies epistemic uncertainty via the difference between predictive entropy and expected entropy:
    \[
    \mathit{MI} = \mathit{H}_{\text{predictive}} - \mathit{H}_{\text{expected}}
    \]
    A higher mutual information implies more disagreement between samples, indicating model uncertainty.

    \item \textbf{Monte Carlo Dropout Variance} — performs forward passes with dropout enabled and measures the variance in predicted probabilities.

    \item \textbf{Class Prediction Variance} — analyzes variation in the final predicted class label across sampled predictions.

    \item \textbf{Class Probability Variance}, \textbf{Sample Variance} — calculate the spread in output probability distributions across multiple samples.

    \item \textbf{Maximum/Minimum Difference Variance} — compute the range between the highest and lowest predicted probabilities for each class across ensemble outputs.

    \item \textbf{Cosine Similarity of Embeddings} — measures semantic alignment by comparing output embedding vectors across predictions.
\end{itemize}

\subsubsection{Input-Level Sensitivity Methods} refer to techniques that assess uncertainty by perturbing the model input (e.g., prompts or contexts) and measuring the sensitivity of predictions.
Representative methods include:

\begin{itemize}[leftmargin=*]
    \item \textbf{SPUQ (Self-Perturbation Uncertainty Quantification)} — paraphrases the input prompt \(N\) times and measures divergence in model outputs using ROUGE-L similarity:
    \[
    \mathit{SPUQ} = \frac{1}{N} \sum_{i=1}^{N} \text{ROUGE-L}(y_0, y_i) \cdot \text{ROUGE-L}(P_0, P_i)
    \]

    \item \textbf{ICE (Input Clarification Ensembles)} — uses multiple semantically equivalent prompts and aggregates the prediction entropy to quantify uncertainty:
    \[
    \mathit{ICE} = \frac{1}{M} \sum_{m=1}^{M} \mathit{Entropy}(\hat{y}_m)
    \]

    \item \textbf{ICL-Sample} — constructs different few-shot contexts and estimates output entropy to reflect prompt sensitivity. 

\end{itemize}

\subsubsection{Reasoning-Level Analyses} assess uncertainty by analyzing the model's internal reasoning processes instead of input perturbations.

\begin{itemize}[leftmargin=*]
    \item \textbf{UAG (Uncertainty-Aware Attention Gradients)} — quantifies the variance in attention scores across different reasoning paths, capturing focus instability:
    \[
    \mathit{UAG} = \frac{1}{T \cdot L} \sum_{t=1}^{T} \sum_{\ell=1}^{L} \operatorname{Var}_{k=1}^{K}(\alpha_{k}^{(\ell, t)})
    \]
    where $\alpha_{k}^{(\ell, t)}$ is the attention score for token $t$ at layer $\ell$ in path $k$.

    \item \textbf{CoT-UQ (Chain-of-Thought Uncertainty Quantification)} — evaluates uncertainty by aggregating weighted keyword frequencies across $K$ reasoning paths:
    \[
    \mathit{CoT\text{-}UQ} = \frac{1}{K} \sum_{k} \sum_{j} f_{j}^{(k)} \cdot w_{j}^{(k)}
    \]
    where $f_{j}^{(k)}$ and $w_{j}^{(k)}$ are frequency and weight of keyword $j$ in path $k$.

    \item \textbf{TopologyUQ} — models reasoning traces as topological persistence diagrams and compares them via Wasserstein distances to quantify structural divergence.

    \item \textbf{Stable Explanations Confidence} — assesses consistency of model-generated explanations by combining answer probability with entailment strength.
\end{itemize}

\subsubsection{Representation-Based Methods} analyze uncertainty based on intermediate model representations rather than final outputs, revealing early-stage uncertainty. Representative methods in this category include:

\textbf{Logit Lens Entropy} computes entropy over softmax-transformed logits from an intermediate layer:
\[
\mathit{LogitLensEntropy}^{(l)} = - \sum_{i=1}^{C} \sigma(\mathbf{z}^{(l)})_i \log \sigma(\mathbf{z}^{(l)})_i
\]





\section{Usage Example}

We implement \dataset as a Python package. One use case of \dataset is presented as follows:

\lstset{
  basicstyle=\footnotesize\ttfamily, frame=tb,
}
\begin{lstlisting}[language=python]
from uncertainty import Quantifier

# Initialize with model and desired method
uq = Quantifier(model, methods=["mc_dropout_var"])

# Compute uncertainty for an input
score = uq.quantify(code_str)
\end{lstlisting}

In this case, the \texttt{Quantifier} class is initialized with a predictive model and one UQ method: \texttt{mc\_dropout\_var}. The \texttt{quantify} method is then called with a code snippet to compute and return its uncertainty score. For detailed usage instructions and guidance on integrating new methods, please refer to the project website\footnote{https://sites.google.com/view/uncertaintyzoo2025}.




\section{Experiments}

We use \dataset to assess how well existing UQ methods reflect LLMs’ prediction confidence in code vulnerability detection.

\subsection{Experimental Setup}

We quantify the prediction uncertainty of fine-tuned~(using the training set of Devign) CodeBERT and pre-trained ChatGLM3-6B-32K models on the test set of Devign~\cite{zhou2019devign}. For UQ methods that require the sampling process~(e.g., MC Dropout Variance, SPUQ, CoT-UQ), we set the sample size as 10. To measure the effectiveness of each UQ method, we calculate the Pearson correlation between the uncertainty scores and prediction correctness. Prediction correctness is computed as a binary indicator (1 for correct, 0 for incorrect) by comparing model predictions with ground-truth labels. Normally, a more significant correlation indicates a more suitable metric for uncertainty quantification. We repeat all experiments five times and report the average results. All experiments were conducted on a Linux system with an NVIDIA H100 GPU, using PyTorch 2.7.1 and HuggingFace Transformers 4.32.1.

\begin{table}[h]
\centering
\caption{Pearson correlation between uncertainty scores and prediction correctness. Input-level and Reasoning-level cannot perform on CoderBERT because it does not support generative tasks. }
\label{tab:correlation_results}
\small
\resizebox{0.45\textwidth}{!}{
\begin{tabular}{lcccc}
\hline
\multirow{2}{*}{\textbf{Method}} & \multicolumn{2}{c}{\textbf{CodeBERT}} & \multicolumn{2}{c}{\textbf{ChatGLM3}} \\
\cline{2-5}
& Pearson Corr. & p-value & Pearson Corr. & p-value \\
\hline
\multicolumn{5}{c}{\textit{Predictive Distribution Methods}} \\
\hline
Avg Neg Log-Likelihood & 0.829 & 0.0000 & 0.293 & 0.0214\\
Avg Probability        & -0.953 & 0.0000 & 0.295 & 0.0389\\
Perplexity             & 0.506 & 0.0000 & 0.207& 0.0781\\
Max Token Entropy      & 0.127 & 0.0000 & 0.185 & 0.0913\\
Avg Pred Entropy       & 0.127 & 0.0000 & 0.185 & 0.0913\\
Token Impossibility Score & 0.953 & 0.0000 & 0.295 & 0.0389\\
Margin                 & -0.106 & 0.0003 & -0.111 & 0.1892\\
Max Probability        & -0.106 & 0.0003 & -0.111 & 0.1892\\
Least Confidence       & 0.106 & 0.0003 & 0.111 & 0.1892\\
Predictive Entropy     & 0.127 & 0.0000 & 0.185 & 0.0913\\
DeepGini               & 0.118 & 0.0000 & 0.180 & 0.0423\\
\hline
\multicolumn{5}{c}{\textit{Ensemble-Based Methods}} \\
\hline
Expected Entropy       & 0.084 & 0.0040 & 0.058 & 0.0849\\
Mutual Info (BALD)     & -0.004 & 0.9009 & -0.007 & 0.8976\\
MC Dropout Var         & 0.049 & 0.0933 & 0.045 & 0.2483\\
Class Pred Var         & 0.027 & 0.3575 & 0.018 & 0.4105\\
Class Prob Var         & 0.066 & 0.0230 & 0.029 & 0.0591 \\
Sample Var             & -0.053 & 0.0703 & -0.020& 0.2906\\
Max Diff Var           & 0.031 & 0.2888 & 0.091 & 0.3890\\
Min Var                & 0.031 & 0.2888 & 0.091 & 0.3890\\
Cosine Sim (Embed)     & -0.036 & 0.2182 & -0.010 & 0.3456\\
\hline
\multicolumn{5}{c}{\textit{Input-Level Sensitivity Methods}} \\
\hline
SPUQ                   & - & - & 0.137 & 0.1846 \\
ICL-Sample             & - & - & 0.181 & 0.1451 \\
ICE (Clarif. Ens.)     & - & - & 0.128 & 0.1964 \\
\hline
\multicolumn{5}{c}{\textit{Reasoning-Level Methods}} \\
\hline
UAG (Attn Grad)        & - & - & 0.017 & 0.278 \\
CoT-UQ                 & - & - & 0.045 & 0.2451 \\
TouT (Tree-of-Thought) & - & - & 0.078 & 0.4761 \\
TopologyUQ             & - & - & 0.066 & 0.3894 \\
Stable Exp Conf        & - & - & 0.256 & 0.2971 \\
\hline
\multicolumn{5}{c}{\textit{Representation-Based Methods}} \\
\hline
Logit Lens Entropy     & -0.021 & 0.4620 & -0.018 & 0.5795 \\
\hline
\end{tabular}
}
\end{table}

\subsection{Results}

Table~\ref{tab:correlation_results} presents the results. We can see a clear difference in the effectiveness of UQ methods between CodeBERT and ChatGLM3. Predictive distribution methods (e.g., Avg Neg Log-Likelihood) achieve high correlation on CodeBERT~(up to 0.953 with P-value less than 0.03), reflecting well-calibrated token probabilities due to task-specific fine-tuning, but show substantially lower correlation on ChatGLM3~(0.18–0.30), indicating poor alignment between token probabilities and prediction correctness in generative models. Ensemble-based methods (e.g., Mutual Information, MC Dropout Variance) exhibit weak or negligible correlation on both models, suggesting limited applicability without task-specific adaptation. On the other side, Input-level methods and reasoning-level methods showcase only modest correlation on ChatGLM3~(0.13–0.26 with no significance), while representation-based methods (Logit Lens Entropy) are largely ineffective, indicating that these approaches provide minimal uncertainty signal in this classification setting. The poor performance of ensemble methods suggests that prediction variance alone may not reliably indicate uncertainty in this task. For generative models like ChatGLM3, the disconnect between token probabilities and task accuracy underscores the challenge of UQ in uncalibrated models.

Overall, these results highlight that token-probability-based UQ methods are effective for fine-tuned classification models like CodeBERT, whereas generative models such as ChatGLM3 require more carefully designed UQ strategies for reliable uncertainty estimation.


\vspace{-5pt}
\section{Conclusion}

We introduce \dataset, a unified and extensible framework for UQ in LLMs, covering 29 UQ methods across both classification and non-classification tasks. 
We demonstrate its application in code vulnerability detection tasks. Experimental results show that many UQ methods effectively capture prediction uncertainty, though their performance varies depending on the method and task type.

\bibliographystyle{ACM-Reference-Format}
\bibliography{reference}

\end{document}